\documentclass[final,twocolumn,times,authoryear]{elsarticle}

\usepackage{framed,multirow}

\usepackage[breaklinks,colorlinks]{hyperref}

\usepackage{pifont}
\usepackage{amsmath,amssymb,amsfonts,amsthm}
\usepackage{algorithm}
\usepackage{algorithmic}
\usepackage{textcomp}
\usepackage{multirow}
\usepackage{hhline, booktabs}
\usepackage{xcolor}
\usepackage{colortbl}
\usepackage{float} 
\usepackage{caption} 
\usepackage{url}
\usepackage{enumerate}
\usepackage{comment}
\usepackage{tikz}
\usepackage{enumitem}
\usepackage{mathabx}
\usepackage{adjustbox}
\usepackage{svg}
\setlist[itemize]{left=0pt, itemsep=0pt, topsep=0pt}

\theoremstyle{plain}
\newtheorem{theorem}{Theorem}[section]

\theoremstyle{definition}
\newtheorem{definition}[theorem]{Definition}

\theoremstyle{remark}

\definecolor{lightgray}{RGB}{243, 247, 245}
\definecolor{ForestGreen}{RGB}{34,139,34}
\definecolor{OrangeRed}{RGB}{236,83,83}
\definecolor{muted-text}{RGB}{240,180,180}

\journal{Pattern Recognition Letters}

\begin{document}

\begin{frontmatter}

\title{Unsupervised contrastive analysis for anomaly detection in brain MRIs via conditional diffusion models}

\author[ubi,inesctec]{Cristiano Patrício\corref{cor1}} 
\cortext[cor1]{Corresponding author.}
\ead{cristiano.patricio@ubi.pt}

\author[unito]{Carlo Alberto Barbano}
\ead{carlo.barbano@unito.it}

\author[unito]{Attilio Fiandrotti} 
\ead{attilio.fiandrotti@unito.it}

\author[unito]{Riccardo Renzulli}
\ead{riccardo.renzulli@unito.it}

\author[unito]{Marco Grangetto} 
\ead{marco.grangetto@unito.it}

\author[feup,inesctec]{Luís F. Teixeira} 
\ead{luisft@fe.up.pt}

\author[ubi]{João C. Neves}
\ead{jcneves@ubi.pt}

\affiliation[ubi]{
    organization={Universidade da Beira Interior and NOVA LINCS},
    city={Covilhã},
    country={Portugal}}

\affiliation[feup]{
    organization={Faculdade de Engenharia da Universidade do Porto},
    city={Porto},
    country={Portugal}}

\affiliation[unito]{
            organization={University of Turin},
            city={Turin},
            country={Italy}}

\affiliation[inesctec]{
    organization={INESC TEC},
    city={Porto},
    country={Portugal}}

\begin{abstract}
Contrastive Analysis (CA) detects anomalies by contrasting patterns unique to a target group (e.g., unhealthy subjects) from those in a background group (e.g., healthy subjects). In the context of brain MRIs, existing CA approaches rely on supervised contrastive learning or variational autoencoders (VAEs) using both healthy and unhealthy data, but such reliance on target samples is challenging in clinical settings. Unsupervised Anomaly Detection (UAD) offers an alternative by learning a reference representation of healthy anatomy without the need for target samples. Deviations from this reference distribution can indicate potential anomalies. In this context, diffusion models have been increasingly adopted in UAD due to their superior performance in image generation compared to VAEs. Nonetheless, precisely reconstructing the anatomy of the brain remains a challenge.
In this work, we propose an unsupervised framework to improve the reconstruction quality by training a self-supervised contrastive encoder on healthy images to extract meaningful anatomical features. These features are used to condition a diffusion model to reconstruct the healthy appearance of a given image, enabling interpretable anomaly localization via pixel-wise comparison.
We validate our approach through a proof-of-concept on a facial image dataset and further demonstrate its effectiveness on four brain MRI datasets, achieving state-of-the-art anomaly localization performance on the NOVA benchmark.
\end{abstract}


\end{frontmatter}


\section{Introduction}
\label{sec:intro}

\begin{figure}[!htb]
\begin{center}
\includegraphics[width=\linewidth]{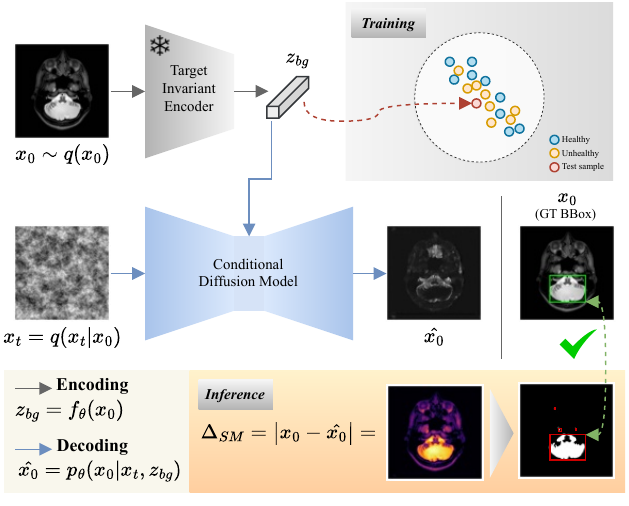}
\caption{\textbf{Overview of the proposed contrastive-guided conditional diffusion model}. The target-invariant encoder $f_\theta(.)$ receives the input image $x_0$ to encode the latent representation $z_{bg}$, which captures common information. The decoder $p_\theta(x_0|x_t, z_{bg})$ receives the noisy sample $x_t$ jointly with the common feature $z_{bg}$ to reconstruct the normal version of the input image. The saliency map is obtained by  $\Delta_{SM} = \left | x_0 - \hat{x_0} \right |$, where high intensity values indicate the anomaly patterns. Intuitively, the class of the image can be determined based on the magnitude of $\bar{\Delta_{SM}}$.}
\label{fig:our-approach}
\end{center}
\end{figure}

Despite substantial progress in image classification models, they still face two significant challenges: i) dependence on extensive amounts of labeled data; and ii) limited interpretability. These challenges are particularly critical in sectors such as healthcare, where explaining the decision process is essential for clinicians to trust the model outcome, and unhealthy samples are scarce and hard to obtain~\citep{patricio2023explainable}. 
Contrastive Analysis (CA) represents a promising solution to this problem~\citep{zou2013contrastive,abid2018exploring,weinberger2022moment} by learning the fundamental generative factors that distinguish a target (TG) dataset from a background (BG) dataset (referred to as anomalous patterns) and are shared across both datasets (referred to as common patterns). However, state-of-the-art CA methodologies have several limitations: i) they often result in blurred images when utilizing Variational Autoencoders (VAEs) or are susceptible to mode collapse and unstable training when using Generative Adversarial Networks (GANs)~\citep{schlegl2019f, carton2024double}; ii) they depend on the availability of healthy (BG) and unhealthy (TG) samples during training; iii) they struggle to preserve common patterns of the original image during the reconstruction process.
On the other hand, UAD has emerged as an established approach for modeling the distribution of healthy anatomy and identifying deviations from this distribution as potential anomalies, thus removing the dependence on annotated unhealthy (TG) samples. While Denoising Diffusion Probabilistic Models (DDPM)~\citep{ho2020denoising} are commonly adopted as generative models in UAD, their training process lacks explicit constraints to preserve common patterns during reconstruction, hindering the fidelity of normal (healthy) reconstructions. Throughout this paper, we use the terms CA and UAD interchangeably, when referring to the task of detecting anomalous patterns in images. 

In this work, we propose an unsupervised methodology for the task of anomaly detection in brain MRIs. Specifically, we build upon a contrastive-guided conditional diffusion model (Figure \ref{fig:our-approach}) to reconstruct a healthy version of the input image while preserving its common anatomical patterns. This is achieved by conditioning the denoising diffusion process on a target-invariant latent feature. To this end, we employ a self-supervised contrastive encoder (referred to as target-invariant encoder) trained exclusively on samples from the healthy (BG) dataset. To approximate potential anomalous variation factors, we apply data augmentation techniques such as random cutout and Gaussian noise, which simulate features like tumor presence in brain MRIs.
During inference, the diffusion model processes unseen samples belonging to the healthy (BG) or unhealthy (TG) distributions by reconstructing the input image such that the anomalous patterns are replaced with learned common (healthy) patterns. By computing pixel-wise differences between the original image and its generated counterpart, high-intensity values can be used to i) classify the input image as {background} (normal) or target (anomalous), based on the magnitude of the reconstruction error; and ii) localize potential anomalies, as illustrated in Figure \ref{fig:our-approach} (bottom-right).

Proof-of-concept validation on facial imagery data demonstrates the utility of our method. Additionally, an extensive comparative analysis against state-of-the-art methods for UAD on four brain MRI benchmarks highlights the effectiveness of our approach, achieving top performance on the recent and challenging evaluation-only NOVA benchmark~\citep{bercea2025nova}. Our major contributions are as follows: 
\begin{itemize}
    \item We propose a methodology for learning common features between {background} (healthy) and {target} (unhealthy) distributions, enabling the generation of a healthy (normal) counterpart of the input image by encoding only common information;
    \item We introduce a target-invariant encoder that learns representations which are both class-invariant and instance-aligned, facilitating common features preservation during the generative process;
    \item We conduct a comprehensive evaluation of our approach on a facial imagery dataset and four brain MRI datasets, encompassing both healthy ({background}) and tumor ({target}) images, to validate the generalization of our approach;
    \item Our method achieves state-of-the-art performance in anomaly localization on the challenging NOVA benchmark.
\end{itemize}

\section{Related Work}
\label{sec:related_work}

Early work in contrastive analysis (CA) relied on the use of Variational Autoencoders (VAEs)~\citep{baur2021autoencoders, behrendt2022capturing, louiset2023sepvae}. A recent method proposed by~\cite{louiset2023sepvae}, SepVAE, aims to distinguish common (healthy) from class-specific (unhealthy) patterns in image data. They utilized VAEs with regularization to encourage disentanglement between common and salient representations, along with a classification loss to separate target and background salient factors. However, the resulting reconstructions were often blurry, limiting their interpretability and utility. Alternative approaches~\citep{schlegl2019f, carton2024double} leveraging Generative Adversarial Networks (GANs) have been explored, but they suffer from issues like mode collapse and unstable training. More recently, Diffusion Denoising Probabilistic Models~\citep{song2021denoising, preechakul2022diffusion} (DDPMs) have emerged as a promising alternative for high-quality image generation, addressing the drawbacks of both GANs and VAEs.
Our work aligns closely with the setting of CA and UAD. In the context of brain MRI, several works have been proposed~\citep{behrendt2023patched, iqbal2023unsupervised, behrendt2025guided}. For example, ~\cite{behrendt2023patched} introduced a patch-based diffusion model for UAD in brain MRIs. They divide the input image into predefined patches and apply noise to each patch individually in the forward pass. In the backward pass, the partly noised image is utilized to recover the noised patch. One drawback of their work is the extensive duration required for inference. More recently, \cite{behrendt2025guided} proposed to conditioning a DDPM to better capture intensity characteristics and domain shifts in brain MRIs during reconstruction, thereby improving segmentation performance. In contrast, our method incorporates contrastive learning during encoder training to explicitly enforce a target-invariant embedding space. 



\section{Background and Notations}
\label{sec:background}

\subsection{Diffusion Models}
\label{subsec:diffusion_models}

Denoising diffusion probabilistic models ~\cite{ho2020denoising} work by corrupting a training image $x_0 \sim q(x_0)$ with a predefined multi-step scheduled noise process to transform it into a sample from a Gaussian distribution. Then, a DNN is trained to revert the process, i.e., starting with a sample from a Gaussian distribution to generate a sample from the data distribution $q(x)$ through a sequence of $T$ sampling steps.

\paragraph{Forward Encoder} Given a training image $x_0$, the noising process consists of gradually noise-corrupting the image $x_0$ by adding Gaussian noise according to some variance schedule given by $\beta_t$:

\begin{equation}
    q(x_t|x_{t-1}) = \mathcal{N}(x_t;\sqrt{1-\beta_t}x_{t-1},\beta_t I).
\end{equation}

As shown in~\cite{song2021denoising}, the noisy version of an image $x_0$ at time $t$ is another Gaussian $q(x_t|x_0) = \mathcal{N}(x_t|\sqrt{\alpha_t}x_0, (1-\alpha_t)I)$ where $\alpha_t = \prod_{s=1}^{t}(1-\beta_s)$, which can be written in the form:

\begin{equation}
\label{eq:xt}
    x_t = \sqrt{\alpha_t}x_0 + \sqrt{1-\alpha_t}\epsilon, \epsilon \sim \mathcal{N}(0,I).
\end{equation}

\paragraph{Reverse Decoder} Since the reverse process $q(x_{t-1}|x_t)$ is intractable, a DNN is used to approximate the distribution $p_\theta(x_{t-1}|x_t)$, where $\theta$ represents the weights and biases of the network. The reverse process is then modeled using a Gaussian distribution of the form:

\begin{equation}
    \label{eq:reverse_process_ddpm}
    p_\theta(x_{t-1}|x_t) = \mathcal{N}(x_{t-1};\epsilon_\theta(x_t,t),\beta_tI),
\end{equation}

where $\epsilon_\theta(x_t,t)$ is a deep neural network governed by a set of parameters $\theta$. From $p_\theta(x_{1:T})$ in Equation \ref{eq:reverse_process_ddpm}, one can generate a sample $x_{t-1}$ from a sample $x_t$ via:

\begin{equation}
    \label{eq:generating_sample_ddpm}
    \resizebox{\columnwidth}{!}{$
    \begin{aligned}
    x_{t-1} = \sqrt{\alpha_t-1}\left(\frac{x_t-\sqrt{1-\alpha_t}\epsilon_\theta(x_t, t)}{\sqrt{\alpha_t}}\right )
    +\sqrt{1-\alpha_{t-1}}\epsilon_\theta(x_t,t)+\sigma_t\epsilon_t,
    \end{aligned}
    $}
\end{equation}

where $\epsilon_t \sim \mathcal{N}(0,I)$ is standard Gaussian noise. For training the decoder, \cite{ho2020denoising} reformulated the variational lower bound objective function and considered the objective of predicting the total noise component added to the original image to create the noisy image at a given step. The loss function is then given by the squared difference between the predicted noise $\epsilon_\theta(x_t, t)$ and the actual noise $\epsilon_t$, for a given time step $t$, using a U-Net~\citep{ronneberger2015u}:

\begin{equation}
    \label{eq:objective_function_original}
    \mathcal{L}_{diff} = \left \| \epsilon_\theta(x_t, t) - \epsilon_t \right \|_2^2.
\end{equation}

By setting $\sigma_t=0$ in Equation \eqref{eq:generating_sample_ddpm}, the coefficient before the random noise $\epsilon_t$ becomes zero, resulting in the following deterministic process of generating a new sample $x_t$. This modification allows to speed up the sampling process without degrading the quality of the generated samples~\citep{bishop2024deeplearning}. 

\subsection{Contrastive Learning}
Contrastive Learning (CL) approaches aim at pulling positive samples' representations (e.g. of the same class) closer together while repelling representations of negative ones (e.g. different classes) apart from each other. Contrasting positive pairs against negative ones is an idea that dates back to previous research~\citep{hadsell_dimensionality_2006, oord_representation_2019, tian_contrastive_2020} and has seen various applications in different tasks, such as face recognition~\citep{schroff2015facenet}. 
Let $x \in \mathcal{X}$ be an \emph{anchor} sample, $x^+$ a positive sample (wrt the anchor), and $x^-$ a negative sample. CL methods look for a parametric mapping function $f:\mathcal{X} \rightarrow \mathbb{S}^{d-1}$ that maps semantically similar samples close together in the representation space (i.e. a hypersphere) and dissimilar samples far away from each other. Once pre-trained, $f$ is fixed, and its representation is evaluated on a downstream task, such as classification, through linear evaluation on a test set. Depending on how positive and negative samples are defined, CL can be employed in self-supervised~\citep{chen_simple_2020} or supervised~\citep{khosla2020supervised} settings.

\section{Method}
\label{sec:methods}

Our method employs a target-invariant encoder to extract a latent representation capturing common features of the input image. A diffusion-based model is then conditioned on this latent representation to reconstruct a normal (healthy) counterpart of the input. The absolute difference between the reconstructed and original images reveals potential anomalies, where brighter areas indicate abnormalities. An overview of the proposed approach is presented in Figure \ref{fig:our-approach}.

\subsection{Target-Invariant Encoder}
\label{subsec:encoder}
The first block of our proposed approach is represented by the target-invariant encoder. This encoder has the goal of learning an input representation which is invariant to the target variable (e.g. presence of eyeglasses in facial images or tumor in brain MRIs) but retains the common information of the input sample. Figure \ref{fig:our-approach} (top-right) presents a visualization of the embedding space after projecting both healthy and unhealthy samples from the trained encoder. The rationale of this approach is that an invariant representation can allow us to correctly reconstruct a realistic normal version of the input image, as it only encodes its common (healthy) information. 
To explain the formulation of our encoder, we introduce three definitions specifying the properties that ensure that the encoder preserves the common information of the image.

\begin{definition}(\emph{Instance-aligned encoder}) 
    Given an anchor $x$, a positive sample $x^+$ of the same subject, and the set of negative samples $x^-_j$ (all other subjects), we say that an encoder $f$ is instance-aligned if:
    \begin{equation}
        \label{eq:instance-alignment}
        ||f(x) - f(x^-_{j})||^2_2 - ||f(x) - f(x^+)||^2_2 \geq \epsilon  \quad \forall j
    \end{equation}
    where $\epsilon \geq 0$. As the margin $\epsilon$ increases, $f$ will provide a better separation between different subjects. In practice, Eq.~\ref{eq:instance-alignment} can be expressed in terms of cosine similarity\footnote{As representations are normalized, i.e. $||f(x)||_2=1$, then $sim(f(x), f(x^+)) = 1 - d(f(x), f(x^+))$ where $d$ is a L2-distance function.}: $sim(f(x), f(x^-_j)) - sim(f(x), f(x^+)) \leq - \epsilon \quad \forall j$  which corresponds to the $\epsilon$-InfoNCE loss~\citep{barbano2023unbiased}:
\begin{equation}
    \label{eq:einfonce}
    \mathcal{L}_{\epsilon-InfoNCE} = - \log \left( \frac{\exp(s^+)}{\exp(s^+ - \epsilon) +  \sum_j \exp(s_j^-)} \right)
\end{equation}
where $s^+$ and $s^-_j$ are shorthand notations for $sim(f(x), f(x^+))$ and $sim(f(x), f(x^-_j))$ respectively. To obtain a sample $x^+$ of the same subject of $x$, if it is not available in the training data, it is possible to employ an augmentation scheme such as in SimCLR~\citep{chen_simple_2020}, i.e. $x^+ = t(x)$ where $t \sim \mathcal{T}$ is an augmentation operator sampled from a family of standard augmentation $\mathcal{T}$ (e.g. random transformations, cropping, etc.).
\end{definition}

\begin{definition}(\emph{Class-invariant encoder})
    \label{def:target_invariant_encoder}
    Denoting with $\mathcal{H} \subset \mathcal{X}$ the set of samples which share the same target attribute value (e.g. healthy), and assuming a binary case for simplicity, we say that $f$ is class-invariant, if $x \in \mathcal{H}$ and $x^+ \in \mathcal{X}\setminus\mathcal{H}$. This means that the alignment in the latent space will be performed between samples with a different target attribute, hence achieving invariance. In a SSL setting, to avoid the dependence on anomalous samples, we leverage data augmentation and data manipulation techniques~\citep{dufumier2023integrating} for approximating the distribution of the target attribute. For example, the appearance of tumors in brain MRIs can be approximated by employing random cutout, or Gaussian noise~\citep{behrendt2023patched} (more details in the supplementary material).
\end{definition}

\begin{definition}(\emph{Target-Invariant Encoder})
    An encoder $f$ preserves common patterns of the image if it is both instance-aligned and class-invariant.
\end{definition}

The above definitions provide the theoretical support of the contrastive learning approach used for training the target-invariant encoder. Considering that the learning process of the encoder is based on a contrastive learning strategy that is instance-aligned and class-invariant, our encoder is expected to produce features which preserve common information of the input image, regardless of whether the image belongs to the background or target set.

\subsection{Conditional Diffusion-based Decoder}
\label{subsec:cdiffmodel}
Our conditional diffusion decoder $p_\theta(x_{t-1}|x_t, z_{bg})$ takes as input the noisy sample $x_t$ and the common feature $z_{bg} \in \mathbb{R}^{1\times d} = f_\theta(x_0)$, a non-spatial vector of dimension $d$ that encodes common patterns observed in the input sample, derived from the properties of the target-invariant encoder $f_\theta$ (Section \ref{subsec:encoder}). Our primary objective is to reconstruct the {background} version of the input image, preserving its common information. Hence, we deviate from the step-wise sampling process in Equation \eqref{eq:reverse_process_ddpm}, which is typically employed to generate new images from noise $p_\theta(x_{T:1}), x_T \sim \mathcal{N}(0,I)$. Instead, we directly estimate the input image $\hat{x_0} \sim p_\theta(x_0|x_t, z_{bg})$ given the noisy sample $x_t$ (more details in Algorithm 2 in the supplementary material). This is achieved by revising Equation \eqref{eq:xt}, following~\citep{song2021denoising}, allowing prediction of the denoised observation, which is an estimation of $x_0$ given $x_t$:

\begin{equation}
    g_\theta(x_t, t, z_{bg}) := (x_t - \sqrt{1-\alpha_t} \cdot \epsilon_\theta(x_t, t, z_{bg})) / \sqrt{\alpha_t}
\end{equation}

The model is then trained (Algorithm 1 in the supplementary material) by minimizing the following objective function~\citep{preechakul2022diffusion}, which is a modified version of the MSE objective in Equation \eqref{eq:objective_function_original}:

\begin{equation}
    \label{eq:objective_function_our_method}
    \mathcal{L}_{diff} = \sum_{t=1}^T \mathbb{E}_{x_0, \epsilon_t} \left[ \left \| \epsilon_\theta(x_t, t, z_{bg}) - \epsilon_t \right \|_2^2 \right]
\end{equation}

We condition the U-Net using adaptive group normalization (AdaGN), which integrates timestep and conditioning embeddings into each residual block via channel-wise scaling and shifting of normalized activations. Further details are provided in the supplementary material.

\section{Experimental Setup}
\label{sec:experiments}
\subsection{Datasets}
\paragraph{\textbf{Facial Images}} We first conduct proof-of-concept experiments on CelebA~\citep{liu2015faceattributes} dataset, containing 202,599 facial images with diverse attributes. We create a subset focusing on subjects wearing eyeglasses and those without accessories, resulting in 15,353 images divided into two distinct classes: 1) \textbf{\textit{Eyeglasses} (EG)}: Images with the 'Eyeglasses' attribute and no other accessory-related attributes, used solely for evaluation as \textit{{target}} images; 2) \textbf{\textit{No Eyeglasses} (NEG)}: Images without the 'Eyeglasses' attribute or any accessory-related attributes, used for training as \textit{{background}} images. Our target task with this dataset is to remove glasses from input images while preserving subject-common information. Facial images are chosen due to their ease of visual interpretation, as identity preservation can be readily assessed during reconstruction. More details on dataset partitions and preprocessing are provided in the supplementary material.

\paragraph{\textbf{Brain MRIs}}
We evaluate our approach in brain MRIs for the task of tumor detection and localization. For a comprehensive comparison, we utilize the {IXI}\footnote{https://brain-development.org/ixi-dataset/} dataset as a {background} (healthy) reference for training, as in previous studies~\citep{behrendt2023patched, behrendt2025guided}. Evaluation is conducted in the Multimodal Brain Tumor Segmentation Challenge 2021 (BraTS21)~\citep{menze2014multimodal, bakas2017advancing, baid2021rsna} and the multiple sclerosis dataset (MSLUB)~\citep{lesjak2018novel}, featuring tumor and Multiple Sclerosis (MS) samples, respectively. Furthermore, we evaluate our model on the recently evaluation-only NOVA~\citep{bercea2025nova} dataset, comprising 906 brain MRI scans spanning 281 rare and diagnostically diverse pathologies. Notably, only T2-weighted images are used from all datasets. More details on data preprocessing partitioning are provided in the supplementary material.

\subsection{Implementation Details}
As target-invariant encoder, we use a ResNet-50 backbone with a fully connected layer and a 128-dimensional output. For brain MRI experiments, we adapt the Spark2D~\citep{tian2023designing} framework to incorporate the $\epsilon$-InfoNCE loss. Notably, the encoder is pre-trained in the healthy training set. At the inference phase, we fine-tune it along with the denoising network.
For the diffusion model, structured simplex noise~\citep{wyatt2022anoddpm} is applied. During training, timesteps $t \in [1, 1000]$ are sampled uniformly, while inference uses a fixed value of $t=500$. The denoising network follows a U-Net architecture. Training runs for up to 1600 epochs on NVIDIA A40 GPUs with Adam optimizer (learning rate of 1e-4, and a batch size of 32). Code is publicly available at \url{https://github.com/CristianoPatricio/unsupervised-contrastive-cond-diff}.

\section{Results}
\label{subsec:results}

\begin{table}[!t]
\caption{\textbf{Reconstruction quality of methods trained on CelebA (training set) and tested on CelebA (test set).} NEG - No Eyeglasses, EG - Eyeglasses. $\left | \Delta  \right | = \left | NEG-EG  \right |$. RE - Random Erasing.}
\label{tab:results_ssim_mse_lpips_celeba_test_set}
\centering
\setlength{\tabcolsep}{3pt}
\resizebox{0.48\textwidth}{!}{%
    \begin{tabular}{lccc|ccc|ccc}
        \toprule
        \multirow{2}{*}{{Model}} & \multicolumn{3}{c}{{\textbf{SSIM} }$ \uparrow$}  & \multicolumn{3}{c}{{\textbf{MSE} }$\downarrow$} & \multicolumn{3}{c}{{\textbf{LPIPS} }$\downarrow$} \\
        & NEG & EG & \scriptsize{$\left | \Delta  \right |$} & NEG & EG & \scriptsize{$\left | \Delta  \right |$} & NEG & EG & \scriptsize{$\left | \Delta  \right |$} \\
        \midrule
        Diff-AE~\citep{preechakul2022diffusion} & $0.6287$ & $0.6079$ & $0.0208$ & ${0.0104}$ & $0.0119$& $0.0015$ & $0.0847$ & ${0.1041}$ & $0.0194$ \\
        SepVAE~\citep{louiset2023sepvae} & $0.5796$ & $0.4773$ & $0.1023$ & $0.0153$ & $0.0268$ & $0.0115$ & $0.2912$ & $0.3322$ & $0.0410$ \\
        DDIM~\citep{dhariwal2021diffusion} & ${0.9690}$ & $0.6568$ & $\textbf{0.3122}$ & $\underline{0.0004}$ & $0.0095$ & $0.0091$ & ${0.0077}$ & $0.1385$ & $\textbf{0.1308}$ \\
        \midrule
        \rowcolor{lightgray}
        \textbf{Ours} \scriptsize{(w/o RE)} & $\underline{0.9756}$ & $0.6737$ & $0.3019$ & $\textbf{0.0003}$ & $0.0097$ & $0.0094$ & $\underline{0.0060}$ & $0.1198$ & $0.1138$ \\
        \rowcolor{lightgray}
        \textbf{Ours} \scriptsize{(w/ RE)} & $\textbf{0.9763}$ & $0.6701$ & $\underline{0.3062}$ & $\textbf{0.0003}$ & $0.0099$ & $\textbf{0.0096}$ & $\textbf{0.0059}$ & $0.1213$ & $\underline{0.1154}$ \\

        \bottomrule
    \end{tabular}%
    }
\end{table}

We begin by validating the effectiveness of our method on the CelebA dataset. Subsequently, we evaluated its performance in anomaly detection and localization tasks, comparing it with established state-of-the-art UAD methods.

\subsection{Proof-of-Concept with CelebA Dataset}
\begin{figure}[!t]
\begin{center}
    \includegraphics[width=0.9\columnwidth]{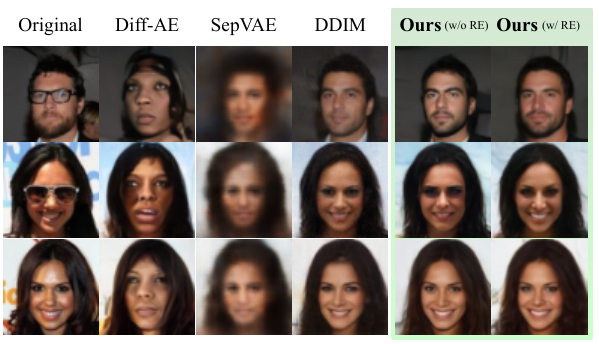}
    \caption{\textbf{Reconstruction results of CelebA images using different methods}. In contrast to Diff-AE, SepVAE and DDIM, our method produces images with well-preserved common information from the input image.}
    \label{fig:recons_celeba}
    \vspace{-20pt}
\end{center}
\end{figure}

Table \ref{tab:results_ssim_mse_lpips_celeba_test_set} presents the reconstruction quality results of our method against baseline models~\citep{preechakul2022diffusion,louiset2023sepvae, dhariwal2021diffusion}. These models were trained in the CelebA dataset using only \textit{{background}} samples (NEG) and evaluated in both \textit{{background}} (NEG) and \textit{{target}} (EG) images.
Evaluation metrics include SSIM ($\uparrow$), LPIPS ($\downarrow$), and MSE ($\downarrow$). Our method (w/ RE) achieves the highest SSIM (0.9763) and the lowest MSE (0.0003) and LPIPS (0.0059) when reconstructing \textit{background} examples (NEG). Although our method (w/o RE) and DDIM produce similar results, they fail to capture common information and high-level details compared to our method (w/ RE), as visually demonstrated in Figure \ref{fig:recons_celeba}.
Furthermore, the supplementary material includes ablation studies analyzing reconstruction quality for different corruption timesteps $t \in [1, 1000]$ and various values of $\epsilon$ in $\epsilon$-InfoNCE.

\paragraph{EG-NEG Classification Performance} Table \ref{tab:results_ssim_mse_lpips_celeba_test_set} provides metrics for distinguishing between the two classes (NEG vs. EG), with the magnitude $\left | \Delta  \right | = \left | NEG-EG  \right |$ indicating distinctiveness (histogram plots are available in the supplementary material). Diff-AE and SepVAE exhibit overlapping distributions across metrics, while our method demonstrates well-separated distributions, yielding the highest ROC AUC score (see Figure C.3 in the supplementary material).

\begin{figure}[!htb]
\begin{center}
    \includegraphics[width=\columnwidth]{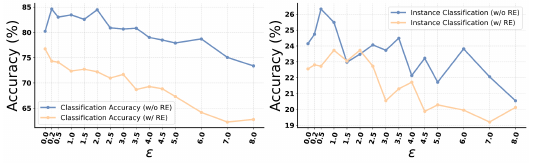}
    \caption{\textbf{Left}: Classification accuracy (NEG vs. EG) for different values of $\epsilon$. \textbf{Right}: Instance classification (accuracy in \%) for different values of $\epsilon$.}
    \label{fig:linear_probing_vs_identity_classification}
\end{center}
\end{figure}

\paragraph{Instance-Alignment}
Varying the margin $\epsilon$ in Equation \eqref{eq:einfonce} can influence identification accuracy by affecting the separation of different classes within the latent space. We conduct an ablation study on $\epsilon$ to observe its impact on identification accuracy using a $k$-NN classifier with $k=3$. Results depicted in Figure \ref{fig:linear_probing_vs_identity_classification} (right) show that the highest accuracy is achieved when $\epsilon$ is chosen from the ranges $[0.5, 2.5]$ with the random erasing (RE) transformation and $[0.15, 1.0]$ without it.

\paragraph{Importance of Class-Invariance} 
We conduct a linear probing analysis to assess the encoder's class-invariance (Definition \ref{def:target_invariant_encoder}). The objective is to evaluate the encoder's capability to distinguish between EG and NEG samples. Subsequently, a logistic regression classifier was trained on the extracted image latent features. Results in Figure \ref{fig:linear_probing_vs_identity_classification} (left) show lower target accuracy with RE, suggesting stronger common information preservation. However, instance classification performs worse with RE, possibly due to built-in augmentations in contrastive learning.

\subsection{Application on Brain MRI Datasets}

Table \ref{tab:models_mri} compares different UAD methods on three brain MRI datasets using evaluation metrics such as DICE score, AUPRC, and $\ell_1$ reconstruction error (calculated only on healthy IXI dataset). We reproduce results for DDPM~\citep{wyatt2022anoddpm}, pDDPM~\citep{behrendt2023patched}, and cDDPM~\citep{behrendt2025guided}, while results for other methods are taken from~\cite{behrendt2023patched}.

\begin{table}[!t]
\centering
\caption{\textbf{Comparison of evaluated models in brain MRI datasets}. Best results are shown in \textbf{bold}, and second-best results are \underline{underlined}. * indicate results obtained by reproducing the method. All metrics are reported as mean $\pm$ standard deviation across 5 folds. Dice score and AUPRC are reported in percentage (\%).}
\label{tab:models_mri}
\setlength{\tabcolsep}{2pt}
\resizebox{0.48\textwidth}{!}{%
\begin{tabular}{lccccccc}
\toprule
\multirow{2}{*}{Model} & \multicolumn{2}{c}{\textbf{BraTS21}} & \multicolumn{2}{c}{\textbf{MSLUB}} & \textbf{IXI} \\
 & DICE $\uparrow$ & AUPRC $\uparrow$ & DICE $\uparrow$ & AUPRC $\uparrow$ & $\ell_1 (1e-3) \downarrow$\\
\midrule

AE\footnotesize{~\citep{baur2021autoencoders}}& $32.87$\scriptsize{$\pm 1.25$} & $31.07$\scriptsize{$\pm 1.75$} & $7.10$\scriptsize{$\pm 0.68$} & $5.58$\scriptsize{$\pm 0.26$} & $30.55$\scriptsize{$\pm 0.27$} \\

VAE\footnotesize{~\citep{baur2021autoencoders}} & $31.11$\scriptsize{$\pm 1.50$} & $28.80$\scriptsize{$\pm 1.92$} & $6.89$\scriptsize{$\pm 0.09$} & $5.00$\scriptsize{$\pm 0.40$} & $31.28$\scriptsize{$\pm 0.71$} \\

SVAE\footnotesize{~\citep{behrendt2022capturing}} & $33.32$\scriptsize{$\pm 0.14$} & $33.14$\scriptsize{$\pm 0.20$} & $5.76$\scriptsize{$\pm 0.44$} & $5.04$\scriptsize{$\pm 0.13$} & $28.08$\scriptsize{$\pm 0.02$} \\

DAE\footnotesize{~\citep{kascenas2022denoising}} & $37.05$\scriptsize{$\pm 1.42$} & ${44.99}$\scriptsize{$\pm 1.72$} & $3.56$\scriptsize{$\pm 0.91$} & $5.35$\scriptsize{$\pm 0.45$} & $10.12$\scriptsize{$\pm 0.26$} \\

f-AnoGAN\footnotesize{~\citep{schlegl2019f}} & $24.16$\scriptsize{$\pm 2.94$} & $22.05$\scriptsize{$\pm 3.05$} & $4.18$\scriptsize{$\pm 1.18$} & $4.01$\scriptsize{$\pm 0.90$} & $45.30$\scriptsize{$\pm 2.98$} \\

DDPM*\footnotesize{~\citep{wyatt2022anoddpm}} & {$39.25$\scriptsize{$\pm 1.01$}} & {$47.79$\scriptsize{$\pm 1.28$}} & 
{$5.43$\scriptsize{$\pm 1.71$}} & 
{$7.25$\scriptsize{$\pm 0.79$}} & 
{$14.10$\scriptsize{$\pm 1.64$}} \\

pDDPM*\footnotesize{~\citep{behrendt2023patched}} & {$49.47$\scriptsize{$\pm 0.91$}} & {$54.68$\scriptsize{$\pm 1.02$}} & 
{$9.17$\scriptsize{$\pm 1.29$}} & 
{$\textbf{10.35}$\scriptsize{$\pm 0.80$}} & 
{$11.31$\scriptsize{$\pm 0.91$}} \\

cDDPM*\footnotesize{~\citep{behrendt2025guided}} & $50.57$\scriptsize{$\pm 1.34$} & $\textbf{56.12}$\scriptsize{$\pm 1.78$} & $\underline{9.88}$\scriptsize{$\pm 0.83$} & $9.45$\scriptsize{$\pm 0.60$} & $\textbf{9.71}$\scriptsize{$\pm 0.47$} \\

\midrule

\rowcolor{lightgray}
\textbf{\textit{Ours}} & $\textbf{50.97}$\scriptsize{$\pm 2.06$} & \underline{$56.11$}\scriptsize{$\pm 2.19$} & $\textbf{10.08}$\scriptsize{$\pm 1.52$} & $\underline{9.65}$\scriptsize{$\pm 1.35$} & \underline{$9.92$}\scriptsize{$\pm 0.59$} \\

\bottomrule
\end{tabular}%
}
\end{table}
\vspace{-10pt}

\paragraph{Reconstruction Quality}
To assess overall reconstruction quality, we use the held-out test set from the healthy IXI dataset and report the $\ell_1$-error, calculated as the average absolute difference between the reconstructed images and their corresponding inputs. As shown in Table \ref{tab:models_mri}, our approach achieves performance competitive with cDDPM and pDDPM, and significantly outperforms all other baselines. These results suggest that the model effectively preserves healthy anatomy from the input image during reconstruction. This can be attributed to the influence of the $\epsilon$-InfoNCE regularization (Eq.~\ref{eq:einfonce}) within the target-invariant encoder, which improves the representation space with regard to the anatomy.

\paragraph{Tumor Segmentation}
To evaluate segmentation performance in the context of UAD, we use all unhealthy test sets and report two metrics: i) Dice score and ii) Area Under Precision–Recall Curve (AUPRC). As shown in Table \ref{tab:models_mri}, our method is competitive with the top-performing baseline, showing marginal improvements in terms of Dice score. We achieve the best results with $\epsilon=0.5$ in Equation \ref{eq:einfonce}. Additional results with other $\epsilon$ values are available in the supplementary material. Analogous to the experimental findings on the CelebA dataset, this model can be used to distinguish between healthy and unhealthy MRIs by analyzing the generated saliency maps, as shown in Figure \ref{fig:our-approach} (bottom-right). For instance, if the difference between a reconstructed brain MRI and its original counterpart shows high-intensity values across large regions, this may indicate presence of tumor and can be flagged for further review by a radiologist. This strategy can be seen as a form of clinical triage, while also contributing to the model's explainability by providing a rationale for its decisions. Additional examples of reconstructions are shown in the supplementary material. 

\begin{figure}[!h]
    \centering
    \includegraphics[width=0.83\columnwidth]{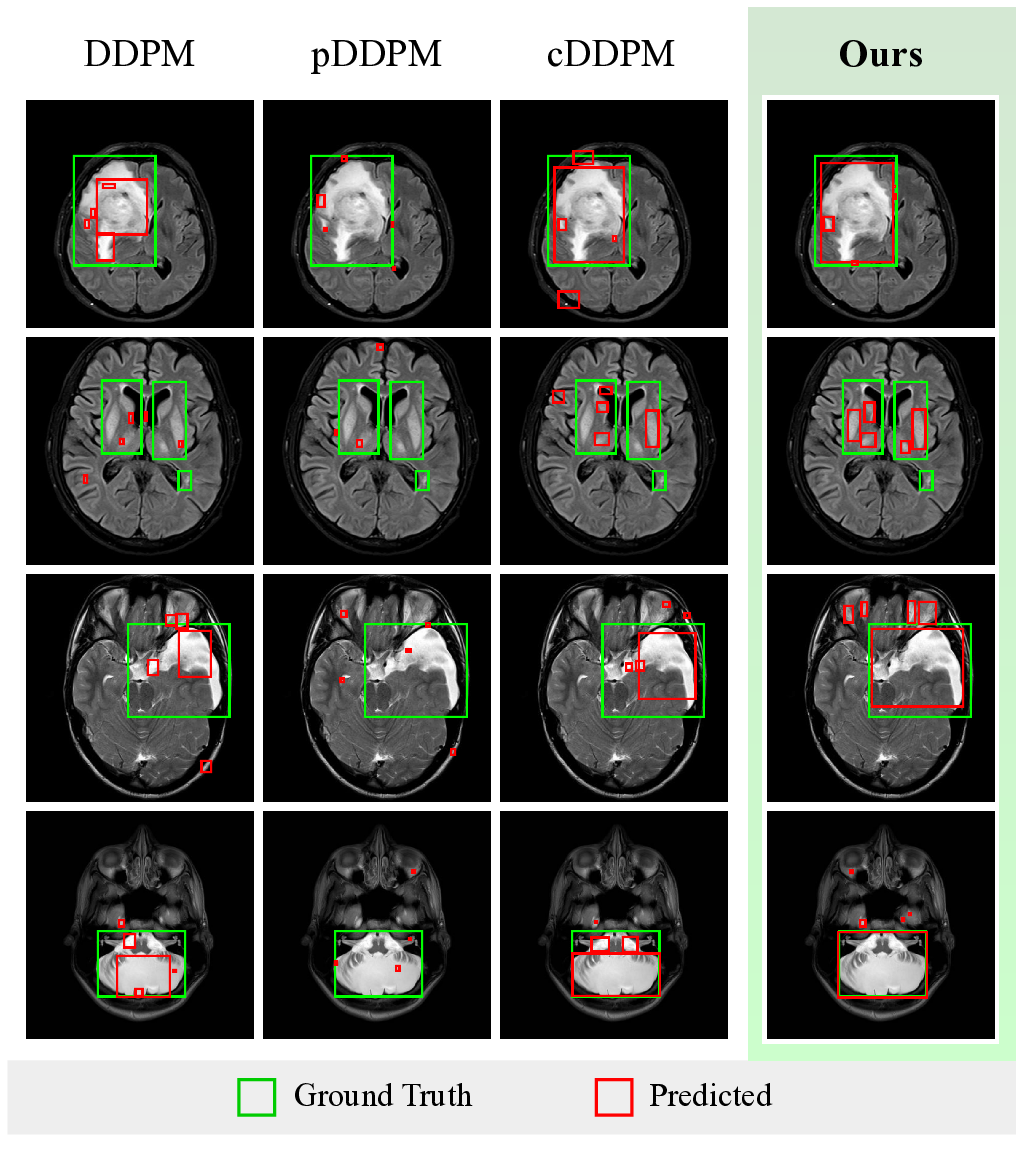}
    \caption{\textbf{Examples of model predictions for anomaly localization on NOVA}. Compared to baseline models, our method produces more precise and spatially focused bounding boxes that better align with ground-truth annotations (green). Each row corresponds to a distinct pathology.}
    \label{fig:qualitative_results_NOVA}
\end{figure}

\paragraph{Anomaly Localization on NOVA Benchmark}
Anomaly localization involves identifying and detect abnormal regions within brain MRIs. This task holds significant clinical importance, as many medical errors arise from failing to detect a pathology altogether. The objective is to predict one or more bounding boxes per image that correspond to abnormal regions, using annotations provided by radiologists as ground truth. 
As shown in Table \ref{tab:results_NOVA}, our method outperforms all baseline approaches in almost all evaluation metrics. Specifically, we achieve these results by considering only T2-weighted images and selecting the top-$5$ predicted largest bounding-boxes. We empirically found that including all modalities available in the NOVA dataset degrades performance, which is expected, as the models were pre-trained only on T2-weighted images from IXI. Compared to pDDPM and cDDPM, our pre-training strategy, which is based on contrastive learning, appears to be more robust than simply using data augmentation. 
As illustrated in Figure \ref{fig:qualitative_results_NOVA}, our model is capable of detecting larger and more accurately localized anomalies, while baseline methods tend to produce smaller and more fragmented predictions. For example, pDDPM underestimates lesion size, likely due to its patching strategy, which may rely on suboptimal patch sizes. These results highlight the effectiveness of our methodology, particularly given that NOVA is considered a challenging and realistic evaluation-only benchmark dataset. Additional ablation studies on modality choice and the selection of top-$N$ largest bounding boxes are provided in the supplementary material.


\begin{table}[!t]
    \centering
    \setlength{\tabcolsep}{3pt}
    \caption{\textbf{Localization performance on NOVA benchmark}. Reported metrics include standard object detection measures (mean Average Precision at multiple IoU thresholds), detection accuracy (ACC50), number of true positives (TP30), and number of false positives (FP30).}
    \resizebox{0.46\textwidth}{!}{%
    \begin{tabular}{lcccccc}
        \toprule
        Model & mAP30 & mAP50 &  mAP50-95 & ACC50 & TP30 & FP30 $\downarrow$ \\
        \midrule 
        DDPM\footnotesize{~\citep{wyatt2022anoddpm}} & 17.50 & 8.23 & 2.75 & 7.36 & 63/394 & 1517 \\
        pDDPM\footnotesize{~\citep{behrendt2023patched}} & 0.30 & 0.30 & 0.06 & 0.25 & 1/394 & 1445 \\
        cDDPM\footnotesize{~\citep{behrendt2025guided}} & 11.51 & 5.19 & 1.51 & 4.72 & 44/394 & \textbf{1423} \\
        
        \midrule
        
        \rowcolor{lightgray}
        \textbf{\textit{Ours}} & \textbf{18.57} & \textbf{9.41} & \textbf{3.33} & \textbf{9.29} & \textbf{71/394} & 1492 \\
        \bottomrule
    \end{tabular}%
    }
    \label{tab:results_NOVA}
\end{table}

\section{Conclusion and Future Work}
\label{sec:conclusion}
In this work, we proposed an unsupervised framework for anomaly detection in brain MRIs that integrates contrastive learning with conditional diffusion models. Our method reconstructs a healthy version of input images while preserving common patterns, enabling both accurate classification and precise and interpretable localization of anomalies. 
By relying exclusively on healthy data, our approach mitigates the reliance on scarce or poorly annotated unhealthy samples. 
Moreover, the use of saliency maps derived from reconstruction errors provides intuitive visual evidence that enhances model interpretability, addressing key challenges in clinical adoption. 
Our results demonstrate the effectiveness of the proposed method, which achieves competitive performance compared to existing UAD methods and ranks first in anomaly localization on the challenging NOVA benchmark.
Nevertheless, our work has limitations and open challenges that are important to acknowledge. The exclusive use of 2D T2-weighted images during training may limit generalization to other structural modalities. 
Future work could explore integrating multimodal imaging data and investigate the model's robustness across diverse pathological conditions. The potential implications of our approach on clinical workflows are discussed in the supplementary material.


\bibliographystyle{model5-names}
\bibliography{refs}

\end{document}